\documentclass[letterpaper, 10 pt, conference]{ieeeconf}  %
\usepackage{siunitx}

\IEEEoverridecommandlockouts                              %

\usepackage[backend=bibtex,bibstyle=ieee,citestyle=numeric-comp,maxbibnames=3,maxcitenames=3,doi=false,url=false,eprint=false]{biblatex}
\usepackage{graphicx}
\usepackage{algpseudocode}
\usepackage{algorithm}
\usepackage{graphics} %
\usepackage{svg}
\usepackage{tabularx}
\usepackage{tablefootnote}

\usepackage{amsmath} %
\usepackage[nameinlink,capitalize]{cleveref}
\Crefname{tabular}{Tab.}{Tabs.}

\usepackage[font=footnotesize]{subcaption}

\title{\LARGE \bf
Boosting LiDAR-Based Localization with Semantic Insight: \\ 
Camera Projection versus Direct LiDAR Segmentation
}

\author{Sven Ochs$^{1}$, Philip Sch\"orner$^{1}$, Marc René Zofka$^{1}$ and J. Marius Z\"ollner$^{1,2}$%
\thanks{$^{1}$ Department of Technical Cognitive Systems, FZI Research Center for Information Technology, 76131 Karlsruhe, Germany.
	{\tt\small ochs, schoerner, zofka, zoellner@fzi.de}}%
\thanks{$^{2}$ Karlsruhe Institute of Technology (KIT), Germany.}
 }

\begin{document}

\maketitle
\thispagestyle{empty}
\pagestyle{empty}

\begin{abstract}

Semantic segmentation of LiDAR data presents considerable challenges, particularly when dealing with diverse sensor types and configurations. However, incorporating semantic information can significantly enhance the accuracy and robustness of LiDAR-based localization techniques for autonomous mobile systems. We propose an approach that integrates semantic camera data with LiDAR segmentation to address this challenge. By projecting LiDAR points into the semantic segmentation space of the camera, our method enhances the precision and reliability of the LiDAR-based localization pipeline.

For validation, we utilize the CoCar NextGen platform from the FZI Research Center for Information Technology, which offers diverse sensor modalities and configurations. The sensor setup of CoCar NextGen enables a thorough analysis of different sensor types. Our evaluation leverages the state-of-the-art Depth-Anything network for camera image segmentation and an adaptive segmentation network for LiDAR segmentation \cite{uecker_one_2025}.
To establish a reliable ground truth for LiDAR-based localization, we make us of a Global Navigation Satellite System (GNSS) solution with Real-Time Kinematic corrections (RTK). Additionally, we conduct an extensive 55 km drive through the city of Karlsruhe, Germany, covering a variety of environments, including urban areas, multi-lane roads, and rural highways. This multimodal approach paves the way for more reliable and precise autonomous navigation systems, particularly in complex real-world environments.

\end{abstract}

\section{INTRODUCTION}
\label{sec:introduction}

Accurate and robust localization is a fundamental requirement for autonomous vehicles in complex environments, such as urban canyons and parking spaces. LiDAR sensors, with their ability to provide high-resolution depth information, play a crucial role in modern localization systems. However, LiDAR-based localization methods often struggle in challenging conditions, such as feature-sparse environments, dynamic scenes, or adverse weather conditions. To address these limitations, semantic information can be integrated into localization frameworks, enhancing the robustness and accuracy of pose estimation.

Recent advancements in machine learning, particularly deep neural networks (DNN), have led to significant improvements in the semantic segmentation of LiDAR point clouds. These methods can classify point cloud data into meaningful categories, providing valuable contextual information for localization. However, LiDAR-only semantic segmentation approaches face inherent challenges, including sensor sparsity, domain adaptation issues, and computational complexity \cite{uecker_one_2025}. 

\begin{figure}[H]
    \centering
    \begin{subfigure}[t]{0.49\columnwidth}
        \centering
        \includegraphics[width=\textwidth]{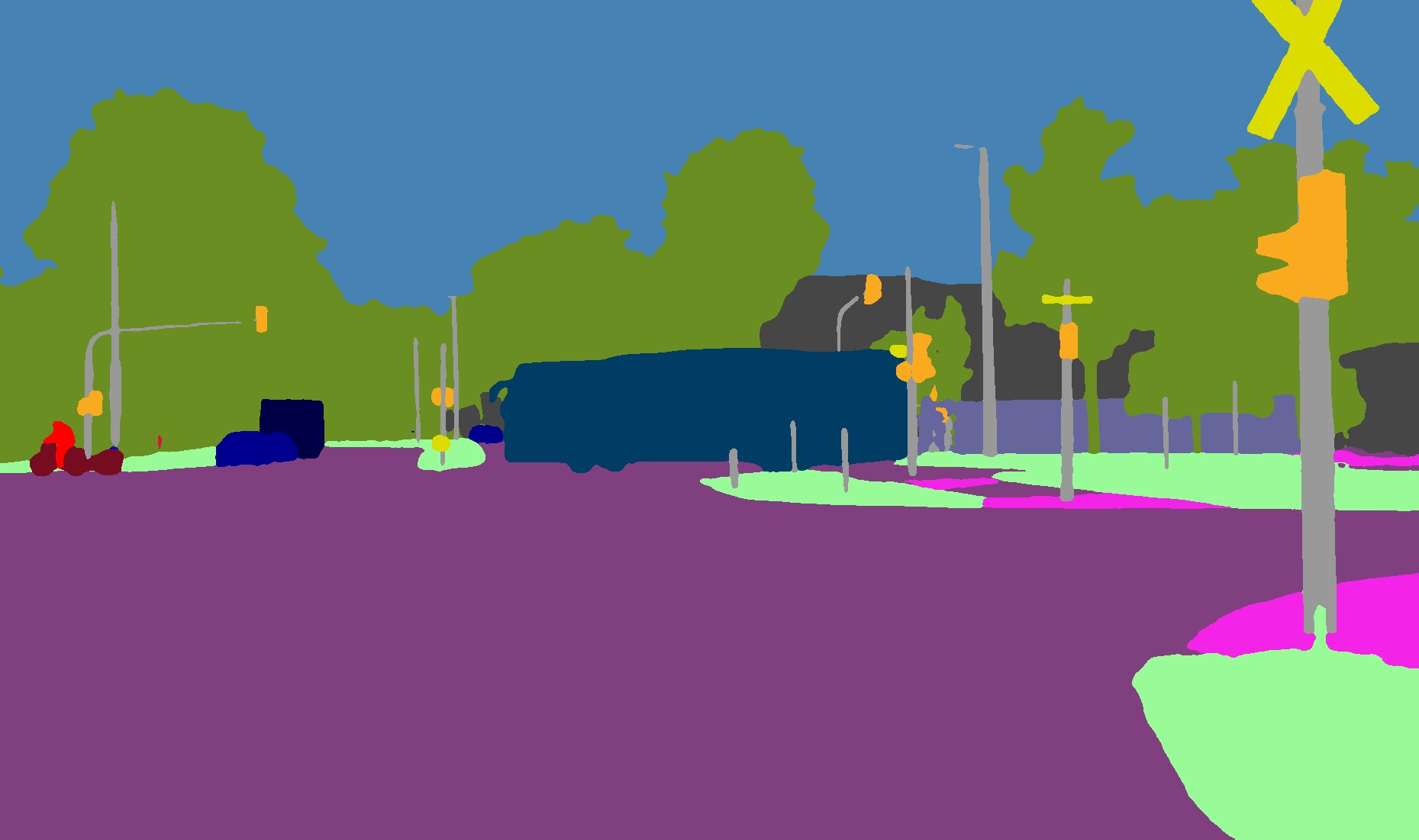}
        \caption{Segmentation of the camera image}
        \label{fig:camera_seg}
    \end{subfigure}
    \begin{subfigure}[t]{0.49\columnwidth}
        \centering
        \includegraphics[width=\textwidth]{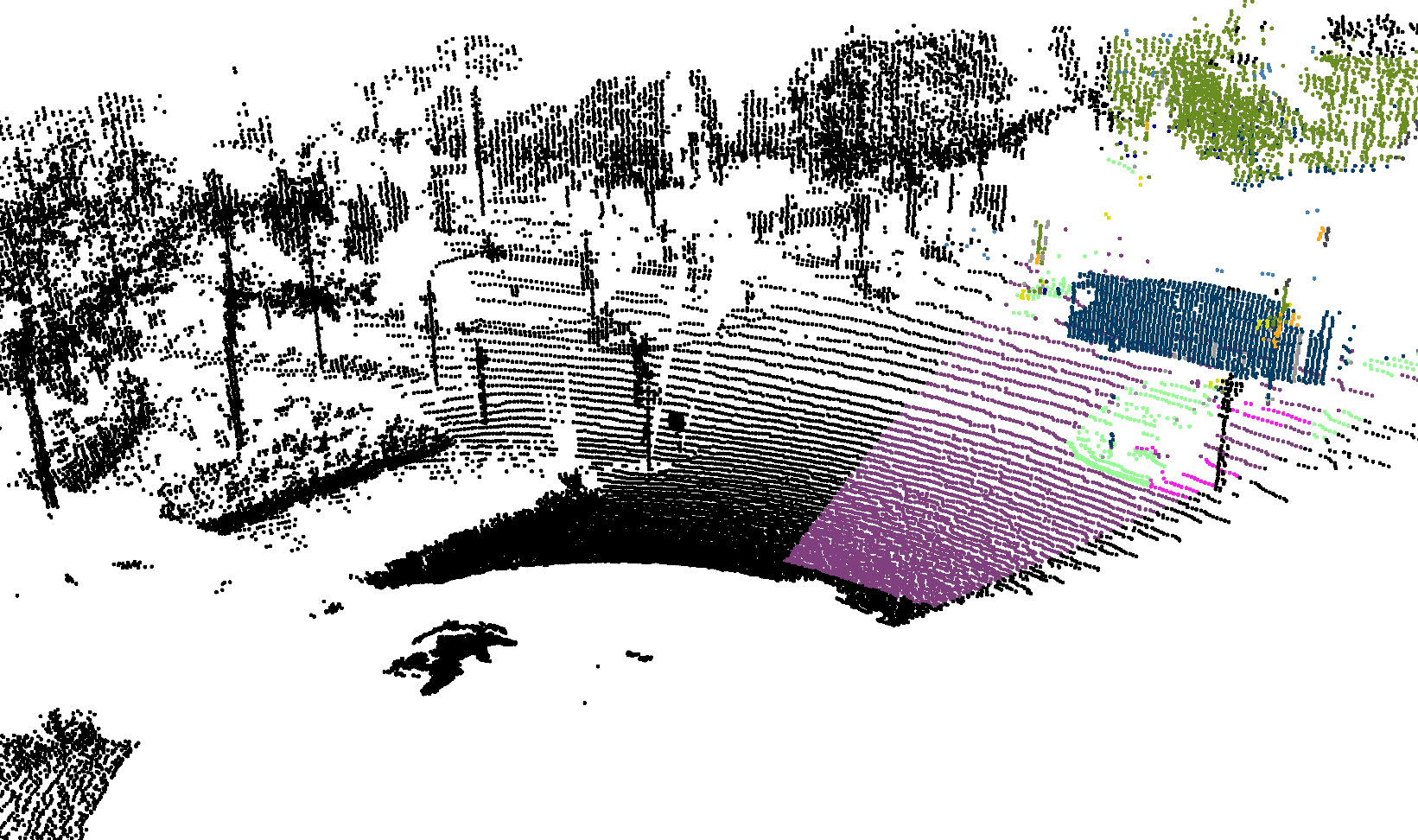}
        \caption{Segmentation through projection into a segmented camera image}
        \label{fig:camera_seg_projection}
    \end{subfigure}
    
    \begin{subfigure}[t]{\columnwidth}
        \centering
        \includegraphics[width=0.49\textwidth]{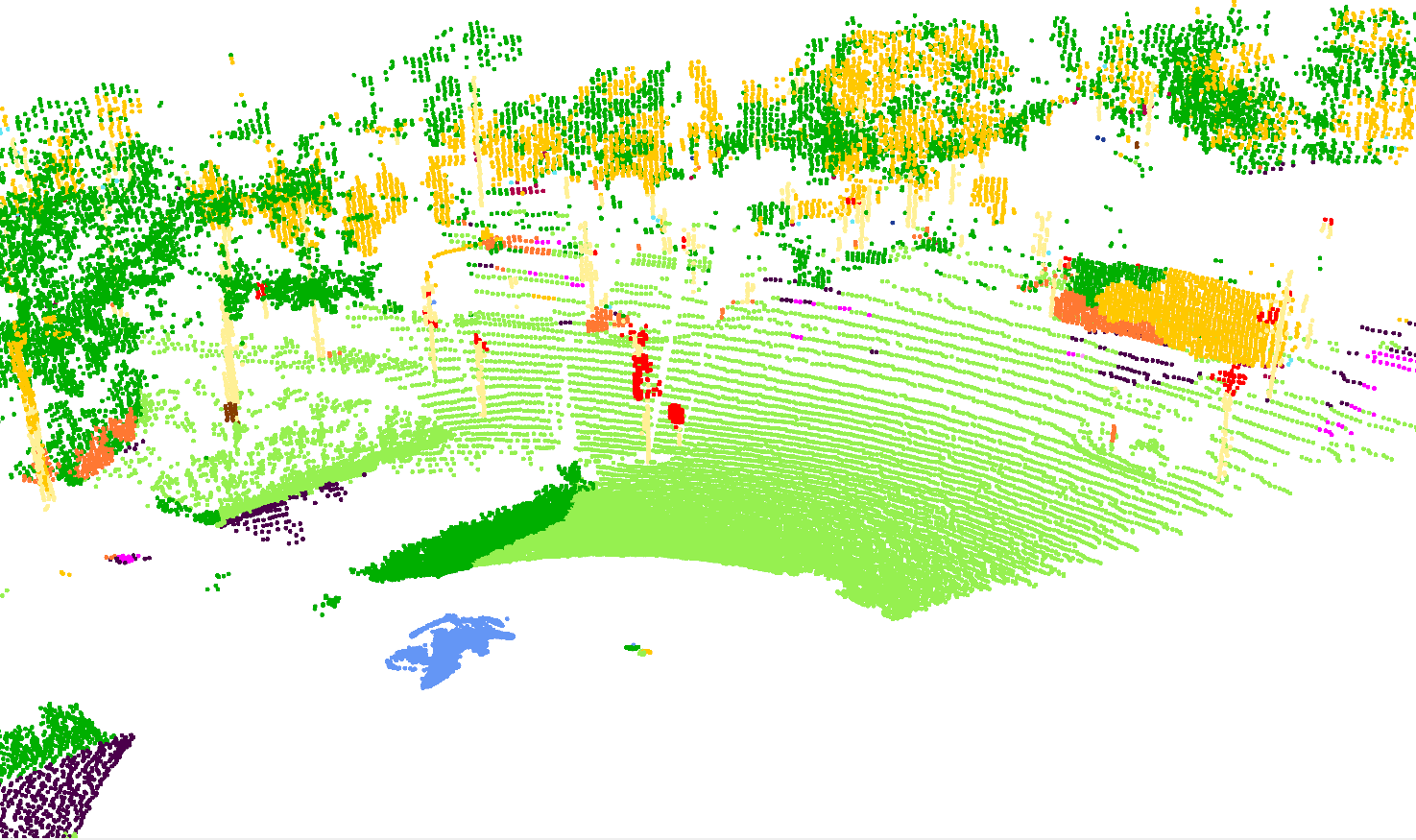}
        \caption{Direct segmentation of whole LiDAR scan}
        \label{fig:direct_lidar_seg}
    \end{subfigure}
    
    \begin{subfigure}[t]{\columnwidth}
        \centering
        \includegraphics[width=\textwidth]{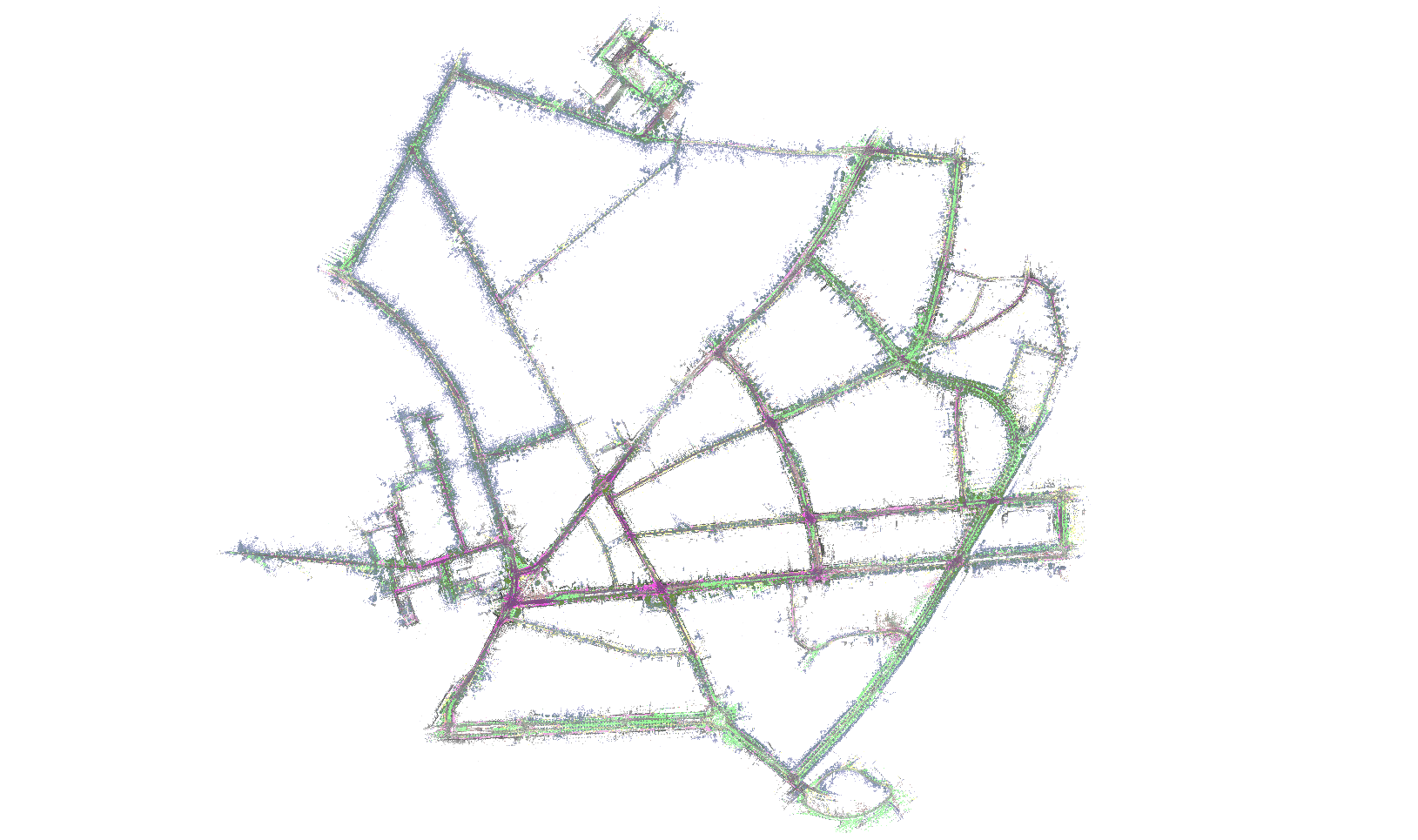}
        \caption{Resulting SLAM-Map utilizing semantic information}
        \label{fig:semantic_mpaping}
    \end{subfigure}
    \caption{Overview of the essential components of our pipeline. \cref{fig:camera_seg} shows the segmentation result from one of the eight cameras. This segmentation is projected onto the LiDAR point cloud in \cref{fig:camera_seg_projection}, and the process is repeated for all cameras to obtain a complete segmentation. \cref{fig:direct_lidar_seg} illustrates the direct segmentation of the LiDAR point cloud using the approach from \cite{uecker_one_2025}. Finally, \cref{fig:semantic_mpaping} presents the semantic maps generated from both segmentation strategies.} 
    \label{fig:control-center}
\end{figure}

In contrast, leveraging semantic information from camera images, it is possible to improve LiDAR-based localization without solely relying on point cloud segmentation. This paper proposes an alternative approach integrating semantic features from camera images into a LiDAR-based localization pipeline through projection techniques. By mapping semantic labels from camera images onto the LiDAR point cloud, we aim to enhance localization performance without requiring direct LiDAR point cloud segmentation.
\newpage
This paper evaluates the proposed method compared to state-of-the-art DNN LiDAR point cloud segmentation techniques. Specifically, we analyze the impact of semantic camera integration on localization accuracy and robustness to environmental variations. Our contributions are as follows:

\begin{itemize}
    \item 
An approach for LiDAR-based localization that integrates semantic camera information through projection.
\item 
A comparative analysis of our method with state-of-the-art machine learning-based semantic segmentation of LiDAR point clouds.
\item 
Experimental validation demonstrates the advantages and disadvantages of localization accuracy.
\end{itemize}

The remainder of this paper is structured as follows: \cref{sec:related_work} provides an overview of related work, including machine learning-based LiDAR segmentation and camera-LiDAR fusion techniques. \cref{sec:method} details the proposed methodology, including the semantic projection pipeline. \cref{sec:evaluation} presents experimental results. Finally, \cref{sec:conclusion} concludes the paper with potential future research directions.

\section{RELATED WORK}
\label{sec:related_work}
\begin{figure*}[t]
    \centering
    \includegraphics[width=\textwidth]{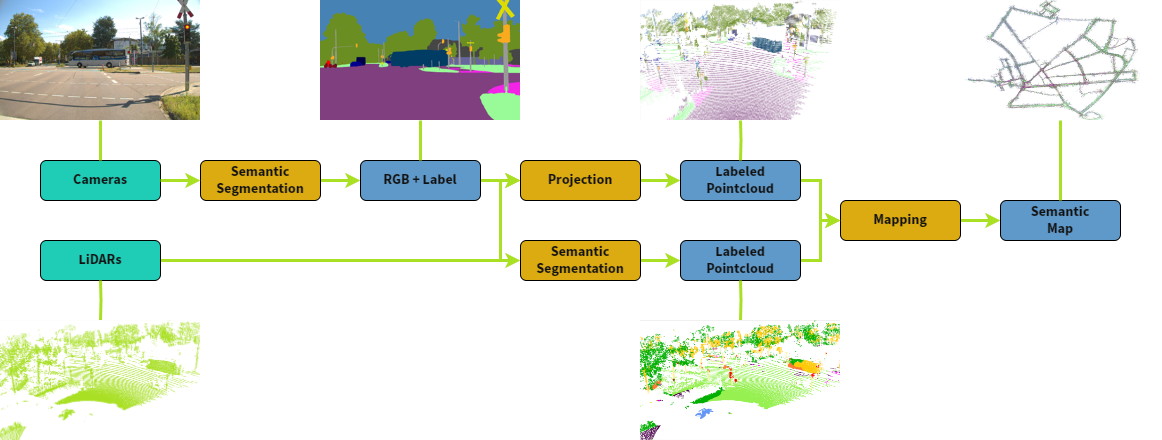}
    \caption{Overview of the pipeline used in our work. The top row illustrates the camera-based segmentation pathway, where the LiDAR point cloud is projected into the image domain to generate a labeled point cloud. The bottom row depicts the direct LiDAR-based semantic segmentation pathway, where the point cloud is segmented without projection. The resulting data is independently used for the same mapping process in both cases.}
    \label{fig:pipeline}
\end{figure*} 

\subsection{LiDAR-Segmentation}

Haidar et al. \cite{haidar_are_2024} evaluate the performance of various 3D semantic segmentation neural networks on resource-constrained NVIDIA Jetson platforms. Utilizing standardized training protocols and data augmentations, the study benchmarks these models using the \emph{SemanticKITTI} and \emph{nuScenes} datasets. Findings indicate that while some models achieve high accuracy, they often fail to meet real-time processing requirements on embedded systems. Conversely, models optimized for speed tend to compromise on segmentation accuracy. This highlights the ongoing challenge of balancing computational efficiency and performance in deploying real-time LiDAR semantic segmentation for autonomous vehicles.

SalsaNext \cite{cortinhal_salsanext_2024} by Cortinhal et al. proposes an enhanced neural network for real-time, uncertainty-aware semantic segmentation of 3D LiDAR point clouds in autonomous driving. It improves upon SalsaNet \cite{aksoy_salsanet_2019} with a Global Context Module, a refined encoder-decoder architecture, and pixel-shuffle upsampling for better segmentation. A combined loss function of weighted cross-entropy and a  Lovász-Softmax optimizes the Jaccard index, improving accuracy. Bayesian uncertainty estimation provides confidence scores for each point’s prediction.

TransRVNet by Cheng et al. \cite{cheng_transrvnet_2023} integrates transformer-based modules like Multi Residual Channel Interaction Attention Module,  Residual Context Aggregation Module, and Balanced Non-square-Transformer Module. These modules effectively capture local and global contextual information within point cloud data. Experimental evaluations demonstrate that TransRVNet performs better than existing state-of-the-art methods on benchmark datasets. Nevertheless, they did not upload the results to the leaderboard of the \emph{SemanticKITTI} semantic segmentation task.

 Yan et al. \cite{yan_2dpass_2022} introduce a training scheme that enhances 3D point cloud semantic segmentation by leveraging 2D image data during training. 2DPASS utilizes an auxiliary modal fusion and multi-scale fusion-to-single knowledge distillation (MSFSKD) approach to transfer rich semantic and structural information from 2D images to a pure 3D network. This enables the model to perform effective segmentation using only point cloud inputs during inference.

Hou et al. \cite{hou_point--voxel_2022} present an approach to improve lightweight LiDAR segmentation models by distilling knowledge from larger, more complex teacher models. This method addresses challenges inherent in point cloud data, such as sparsity and varying density, by transferring knowledge at both the point and voxel levels. Key components include point and voxel-wise output distillation to complement sparse supervision signals and a difficulty-aware sampling strategy emphasizing less frequent classes and distant objects. In addition, the approach employs inter-point and inter-voxel affinity distillation to capture structural information effectively.

\subsection{Image Segmentation}

Segformer by Xie et al. \cite{xie_segformer_2021} combines Transformers with lightweight multilayer perceptron (MLP) decoders for semantic segmentation tasks. SegFormer features a hierarchically structured Transformer encoder that produces multiscale features without relying on positional encoding, enhancing flexibility across varying image resolutions. It's a simple MLP decoder that effectively aggregates information from different layers, integrating local and global attention mechanisms.

Cheng et al. \cite{cheng_masked-attention_2022} present Mask2Former, a versatile architecture capable of addressing various image segmentation tasks, including panoptic, instance, and semantic segmentation. Its core innovation is the masked attention mechanism, which extracts localized features by constraining cross-attention within predicted mask regions. This design eliminates the need for task-specific architectures, streamlining research efforts.

Depth Anything by Yang et al. \cite{yang_depth_2024-1} enhances monocular depth estimation that delivers finer and more robust depth predictions compared to its predecessor \cite{yang_depth_2024}. This improvement is achieved through three key practices: replacing all labeled real images with synthetic images, scaling up the capacity of the teacher model, and employing large-scale pseudo-labeled real images to train student models. Trained on 595 thousand synthetic labeled images and over 62 million real unlabeled images, the model significantly outperforms previous versions in detail and robustness. Additionally, it demonstrates greater efficiency, being more than ten times faster and more accurate than models built on Stable Diffusion.

\subsection{Fusion-based approaches}

Reichardt et al. \cite{reichardt_360circ_2023} present ImageTo360, a method designed to enhance label-efficient LiDAR segmentation by leveraging semantic predictions from a single camera view. This approach utilizes a frozen image-based teacher network to generate semantic labels for LiDAR data within the camera's field of view, which are then used to pre-train a LiDAR segmentation student network. Subsequent fine-tuning on full 360° LiDAR data allows the model to generalize beyond the initial camera view without requiring image data during inference.

 Gu et al. \cite{gu_clft_2024} introduce a vision transformer-based network, CLFT, designed to perform semantic segmentation by fusing camera and LiDAR data. CLFT employs a novel progressive-assemble strategy within a bi-directional network and integrates results using a cross-fusion strategy across transformer decoder layers. Unlike previous methods, CLFT is evaluated under challenging conditions such as rain and low illumination and demonstrates robust performance.

\section{METHOD}
\label{sec:method}

In this section, we would like to present the enhancement of LiDAR data using the projection of semantic camera data. For this, first of all, we present the utilized pipeline, highlighted in \cref{fig:pipeline}. The top row presents the camera projection results, while the bottom row displays the LiDAR-based segmentation.

The camera projection pipeline comprises three fundamental components: semantic segmentation of the camera image, temporal synchronization, and the subsequent projection of LiDAR data into the image plane.
For the semantic segmentation task, we employ the Deep-Anything \cite{yang_depth_2024} pre-trained Vision Transformer Large (ViT-L) encoder in conjunction with downstream semantic segmentation models. This enables the extraction of meaningful scene understanding from the camera images, facilitating accurate object and environmental segmentation. An illustrative example of the semantic segmentation results can be found in \cref{fig:camera_seg}.

Following the segmentation process, precise temporal alignment between the camera images and the LiDAR point cloud data is required to ensure accurate data fusion. This synchronization is achieved through the \emph{Approximate Time Policy} implementation provided by the Robot Operating System (ROS) framework \cite{stanford_artificial_intelligence_laboratory_et_al_robotic_2018}, which mitigates discrepancies in timestamped data streams by determining the most temporally coherent camera-LiDAR pairing.
Once a suitable temporal correspondence between the camera and LiDAR data has been established, each LiDAR measurement point is projected onto the camera image plane. This projection process relies on the sensor system's extrinsic and intrinsic calibration parameters, which define the spatial transformations necessary to map LiDAR coordinates into the camera reference frame.

The parallel processing pipeline dedicated to purely LiDAR-based data, which will be employed in \cref{sec:evaluation}, leverages a network of \cite{uecker_one_2025}. This network is predominantly trained on LiDAR sensors with resolutions of 64×2650 points per scan. However, as outlined in the experimental setup (\cref{sec:testing-vehicle}), our system utilizes a higher-resolution 128×2048 LiDAR sensor.
Despite the absence of fine-tuning specifically for this sensor configuration, the architectural design of \cite{uecker_one_2025} inherently allows it to generalize across different LiDAR resolutions. This adaptability is primarily attributed to its point-voxel representation learning framework, which effectively captures local geometric structures and global spatial contexts. Consequently, even without explicit retraining for the 128×2048 LiDAR configuration, the network can still deliver reasonably robust performance in downstream perception tasks.

Since the camera and LiDAR segmentation are based on different datasets, discrepancies arise in their classifications. The camera segmentation is trained on the Cityscapes dataset \cite{cordts_cityscapes_2016}, while the LiDAR segmentation relies on SemanticKITTI \cite{behley_semantickitti_2019}. In SemanticKITTI, all dynamic classes are remapped to their corresponding static categories—for instance, \textit{moving-car} is reassigned to \textit{car}. Conversely, in Cityscapes, the classes \textit{ego-vehicle}, \textit{rectification border}, \textit{out of ROI}, \textit{static}, \textit{dynamic}, and \textit{sky} are categorized as \textit{unlabeled}.

For the evaluation, we use Chef-Kiss \cite{ochs_chefs_2025}, which extends the capabilities of KISS-ICP by incorporating semantic information into the point alignment process through a generalizable approach with minimal parameter tuning. This enhancement enables Chef-Kiss to surpass KISS-ICP regarding Absolute Trajectory Error (ATE), the key metric for map accuracy. Additionally, Chef-Kiss enhances the Cartographer mapping framework to process semantic information, improving loop closure detection over larger areas, reducing odometry drift, and further refining ATE accuracy. By integrating semantic data into the mapping process, Chef-Kiss enables the filtering of specific object classes, such as parked vehicles, from the final map. This filtering mitigates the effects of temporal changes, such as vehicle displacement, thereby enhancing pure localization quality.

\section{EVALUATION}
\label{sec:evaluation}

\begin{figure*}[ht]
     \begin{subfigure}[t]{0.245\textwidth}
		\includegraphics[width=\textwidth]{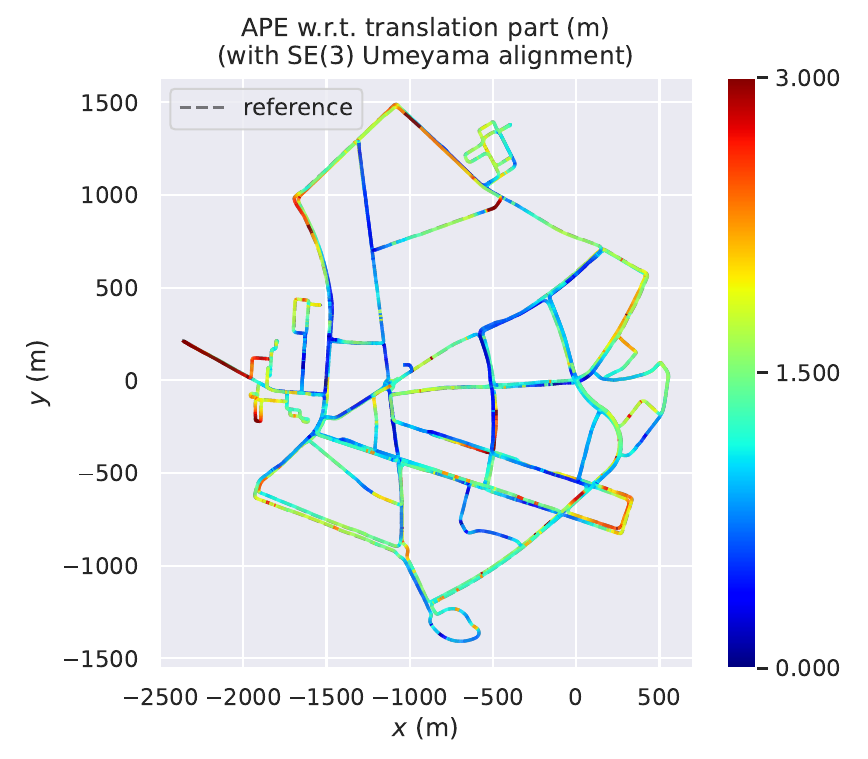}
        \caption{Baseline}
    \end{subfigure}\hfill
    \begin{subfigure}[t]{0.245\textwidth}
		\includegraphics[width=\textwidth]{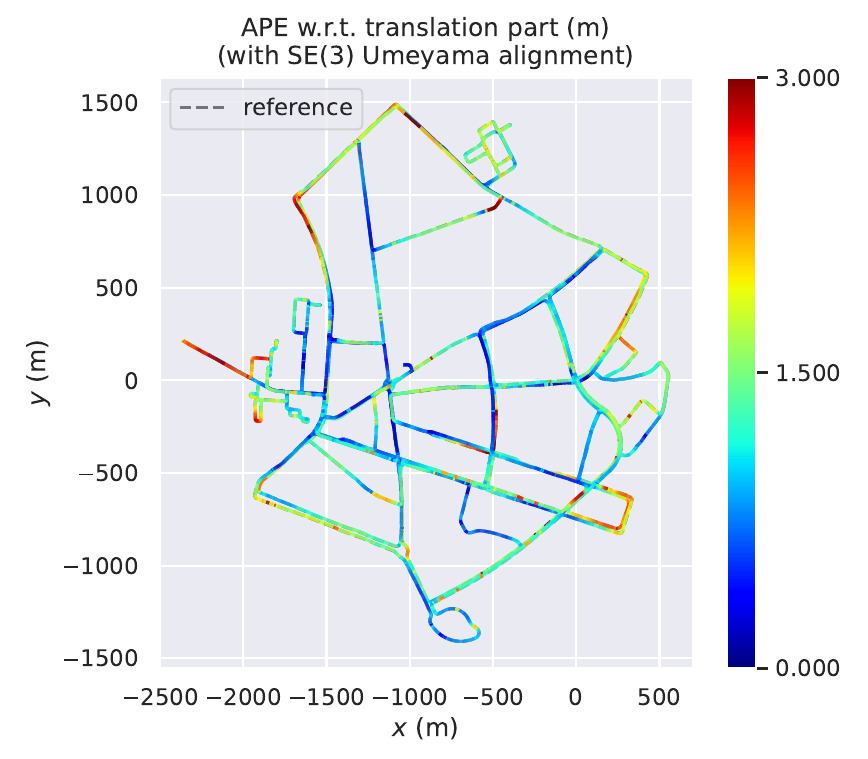}
        \caption{Camera segmentation\\ with ground}
    \end{subfigure}\hfill
    \begin{subfigure}[t]{0.245\textwidth}
		\includegraphics[width=\textwidth]{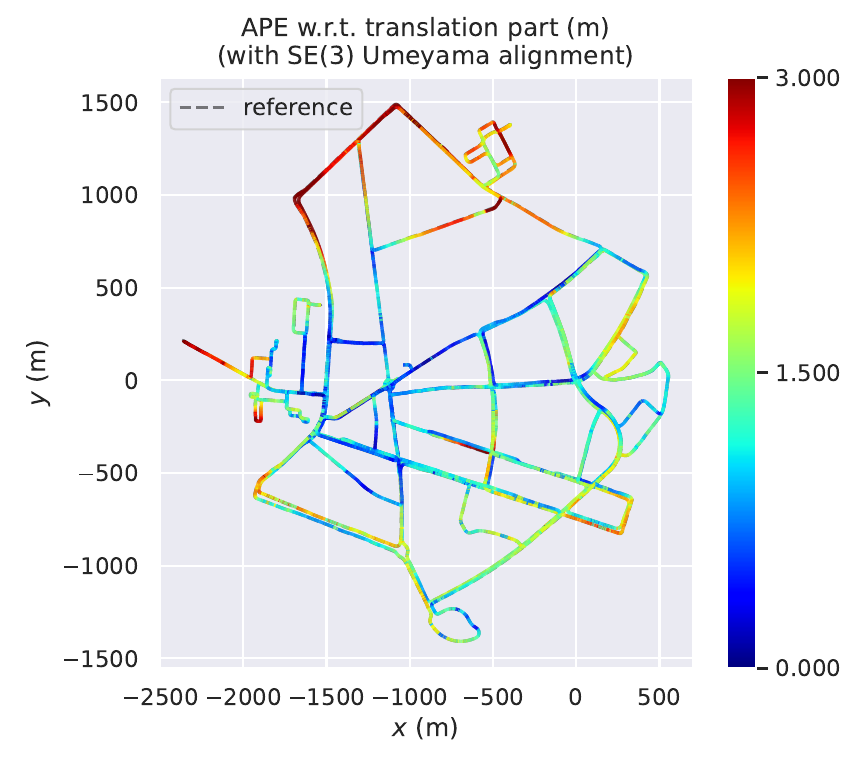}
        \caption{LiDAR segmentation\\ with ground}
    \end{subfigure}\hfill
    \begin{subfigure}[t]{0.245\textwidth}
		\includegraphics[width=\textwidth]{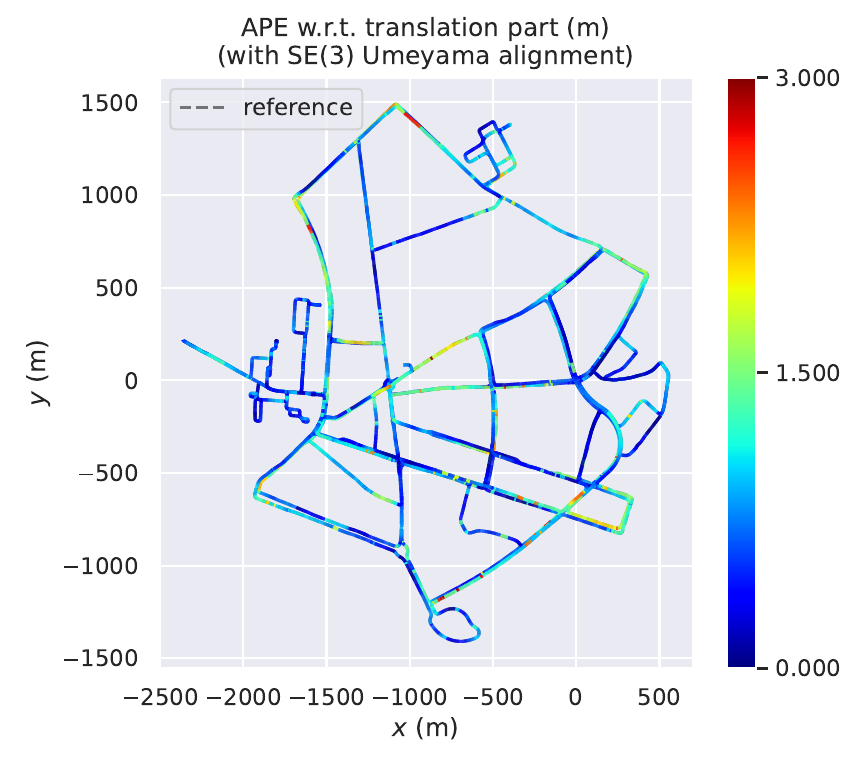}
        \caption{Camera Segmentation\\ with GNSS}
    \end{subfigure}\hfill
    \caption{Trajectories generated using the long-range LiDAR sensor. The trajectory is colored depending on the Euclidean distance to the ground truth. The most prominent errors occur along the road segment in the map's middle left and the upper central part. These errors are primarily attributed to these areas' lack of loop closures, leading to degraded localization performance. A corresponding qualitative analysis is presented in \cref{tab:long_range}. The graphics are generated utilizing \cite{grupp_evo_2017}.}
    \label{fig:long_range_trajectories}
\end{figure*}

\begin{figure*}[ht]  
     \begin{subfigure}[t]{0.245\textwidth}
		\includegraphics[width=\textwidth]{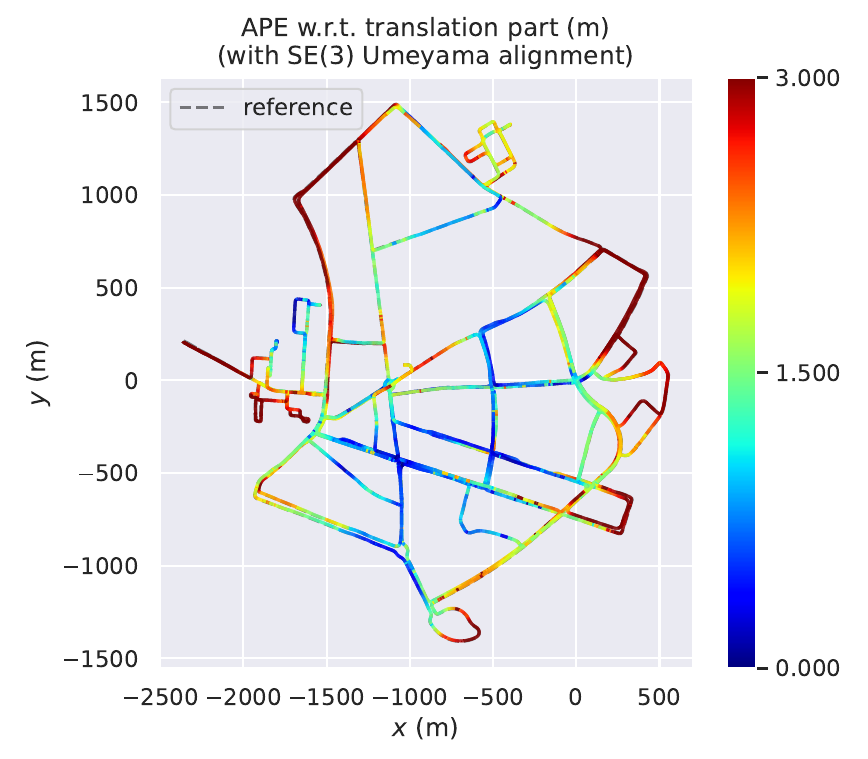}
        \caption{Baseline}
    \end{subfigure}\hfill
    \begin{subfigure}[t]{0.245\textwidth}
		\includegraphics[width=\textwidth]{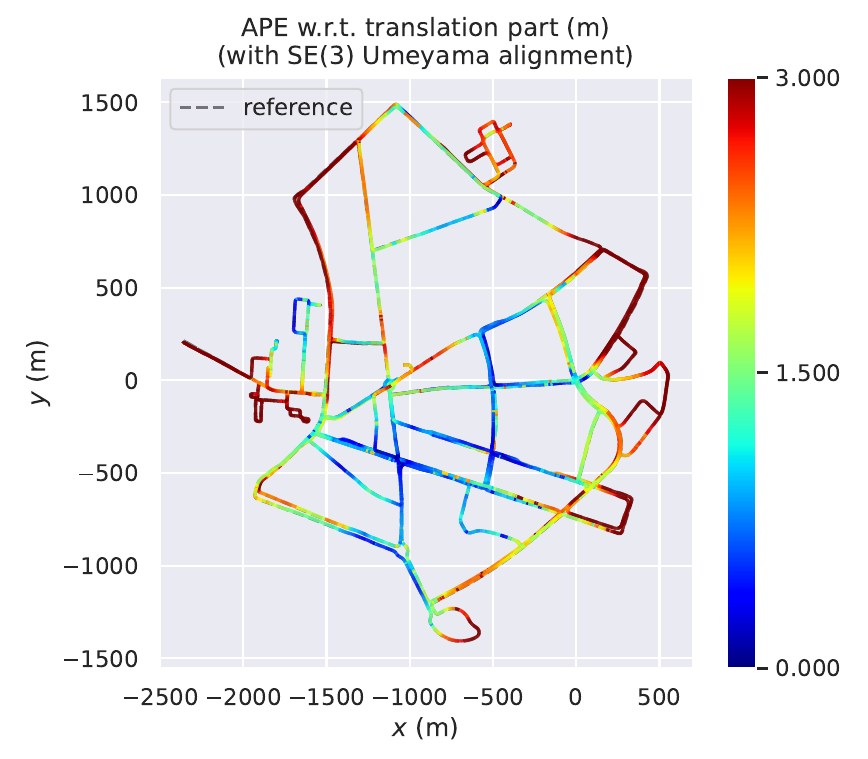}
        \caption{Camera segmentation\\ no ground}
    \end{subfigure}\hfill
    \begin{subfigure}[t]{0.245\textwidth}
		\includegraphics[width=\textwidth]{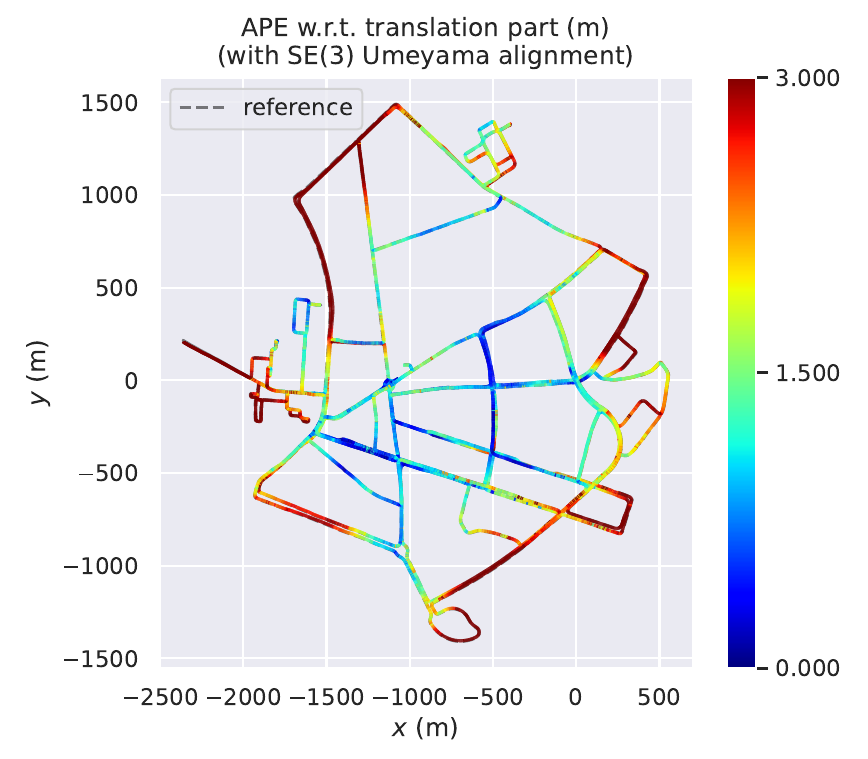}
        \caption{LiDAR segmentation\\ with ground}
    \end{subfigure}\hfill
    \begin{subfigure}[t]{0.245\textwidth}
		\includegraphics[width=\textwidth]{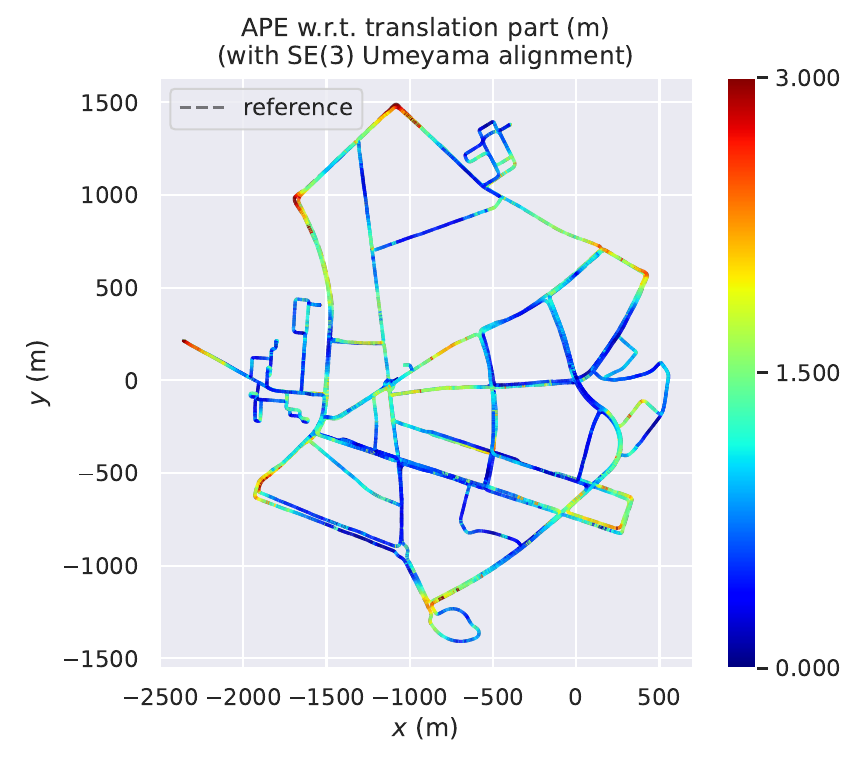}
        \caption{LiDAR segmentation\\ with GNSS}
    \end{subfigure}\hfill
    \caption{Trajectories obtained using the mid-range LiDAR sensor. The trajectory is colored depending on the Euclidean distance to the ground truth. Compared to the long-range LiDAR results shown in \cref{fig:long_range_trajectories}, the mid-range sensor exhibits reduced performance, particularly in less structured areas. However, in feature-rich regions, the system still achieves high-accuracy mapping. The graphics are generated with \cite{grupp_evo_2017}.}
    \label{fig:mid_range_trajectories}
\end{figure*}

This section will comprehensively evaluate the two proposed processing pipelines by assessing their mapping accuracy compared to a high-precision GNSS refined with RTK. The accuracy assessment will be quantified using the Absolute Trajectory Error (ATE), which measures the deviation of the estimated trajectory from the reference GNSS trajectory, thereby providing an objective metric for evaluating the spatial precision of the mapping pipelines.

\subsection{Testing Vehicle}
\label{sec:testing-vehicle}

The dataset utilized for the evaluation was acquired using the testing vehicle CoCar NextGen, provided by the FZI Research Center for Information Technology. This vehicle has an advanced perception system comprising six LiDAR sensors, providing a complete 360-degree horizontal field of view (FOV) with a resolution of 2048 points per scan. Among these, four mid-range LiDAR sensors feature a vertical FOV of 45 degrees, while the remaining two long-range sensors have a vertical FOV of 22.5 degrees, each with a resolution of 128 rows.

In addition to the LiDAR sensors, the vehicle is equipped with eight cameras strategically positioned close to the LiDAR units. The onboard cameras feature diverse fields of view and resolutions, with a specialized configuration at the front of the vehicle that includes telephoto, medium, and wide-angle lenses to optimize perception across different depth ranges and scene contexts. For a more detailed description of the sensor configuration and system specifications, we refer to \cite{heinrich_cocar_2024}.

For the evaluation, all eight cameras are considered to achieve a comprehensive 360-degree surround view, including overlapping front-facing cameras. To handle diverging predictions, previously labeled points are not overwritten. Consequently, the order in which the camera images are processed plays a critical role. The telephoto camera is processed first due to its superior resolution and image quality, followed by the medium and wide-angle cameras. Regarding the LiDAR sensors, we evaluate two different sensor types: the long-range LiDAR and one mid-range LiDAR sensor.

\subsection{Map Generation}

\begin{table}[th]
\begin{tabularx}{\columnwidth}{r|ccccc}
\hline
\textbf{Long-Range-Lidar} & \textbf{max} & \textbf{mean} & \textbf{min} & \textbf{rmse} & \textbf{std} \\ \hline
\textbf{KISS-SLAM}      & 13.79 & 2.56  & 002 & 3.15 & 1.83 \\
\textbf{Baseline}       & 4.53  & 1.47  & 0.03 & 1.63 & 0.70 \\
\textbf{Camera-Non-Ground}     & 4.43  & 1.47  & 0.02 & 1.62 & 0.69 \\
\textbf{Camera-With-Ground}     & 4.03  & 1.24  & 0.01 & 1.37 & 0.58 \\
\textbf{LiDAR-Non-Ground}     & 4.35     & 1.41     & 0.00    & 1.59    & 0.74    \\
\textbf{LiDAR-With-Ground}     & 4.27     &  1.39     & 0.01    & 1.56    & 0.71    \\
\textbf{Baseline-With-GNSS}   & 3.51  & 0.92  & 0.01 & 1.05 & 0.52 \\
\textbf{Camera-With-GNSS}   & 3.20  & 0.80  & 0.00 & 0.93 & 0.47 \\
\textbf{LiDAR-With-GNSS}   & 3.41     & 0.82     & 0.00    & 0.94    & 0.46    \\
\hline
\end{tabularx}
\caption{Quantitative comparison of different localization methods using the long-range LiDAR. Metrics include maximum, mean, minimum, RMSE, and standard deviation of localization error (in meters). Lower values indicate better performance.}
\label{tab:long_range}
\end{table}

\cref{tab:long_range} and \cref{tab:mid-range} presents a comprehensive quantitative comparison of KISS-SLAM \cite{guadagnino_kiss-slam_2025} and Chefs Kiss \cite{ochs_chefs_2025}. The baseline configuration employs Chef's Kiss with the occupancy-based approach, excluding semantic information. The camera rows utilize semantic labels derived from LiDAR via projection, as illustrated in \cref{fig:pipeline}. In contrast, the LiDAR rows apply semantic segmentation directly to the LiDAR data using the method from \cite{uecker_one_2025}. The camera and LiDAR setups are subdivided based on whether ground information is incorporated.

Five statistical metrics are employed to evaluate performance: maximum error, mean error, minimum error, root mean square error (RMSE), and standard deviation. The results demonstrate substantial differences in performance across the evaluated methods, underscoring the influence of sensor modality, ground segmentation, and GPS integration on localization accuracy.

KISS-SLAM, a popular SLAM-based approach, exhibits the most dramatic improvement using long-range LiDAR. In the mid-range setting, KISS-SLAM suffers from high error values, with a mean error of \SI{5.05}{\meter} and an RMSE of \SI{5.95}{\meter}—far exceeding the other methods. However, with long-range LiDAR, its performance improves significantly: the mean error drops to \SI{2.56}{meters}  and the RMSE to \SI{3.15}{\meter}. This suggests that KISS-SLAM heavily benefits from a broader sensing range, although it still lags behind simpler or GNSS-assisted approaches.

The 'Baseline' method without sensor fusion performs well across both sensor configurations. In the mid-range scenario, it achieves a mean error of \SI{1.80}{\meter} meters and an RMSE of \SI{2.12}{\meter}, while in the long-range case, these values improve to \SI{1.47}{\meter} and \SI{1.63}{\meter}, respectively.

When considering methods combining camera data without GNSS, 'Camera-Non-Ground' and 'Camera-With-Ground' benefit from long-range sensing. The ground-filtered variant, in particular, shows substantial improvement, with its mean error decreasing from \SI{1.95}{\meter} to \SI{1.24}{\meter} and its RMSE from \SI{2.35}{\meter} to \SI{1.37}{\meter}. This pattern also applies to LiDAR-based methods without GNSS. 'LiDAR-With-Ground' and 'LiDAR-Non-Ground' both see a reduction in mean error from around \SI{1.91}{\meter} and \SI{1.96}{\meter} to \SI{1.39}{\meter} and \SI{1.41}{\meter} and improved RMSE values, underscoring that extended range and ground-aware processing contribute to better localization.

The most accurate results, however, come from methods that incorporate GNSS. Regardless of the sensing modality—whether camera, LiDAR, or the baseline approach—GNSS-assisted localization consistently achieves the lowest error across all metrics. For example, 'LiDAR-With-GNSS' achieves a mean error of \SI{0.82}{\meter} and an RMSE of \SI{0.94}{\meter} in the long-range setup, compared to \SI{0.99}{\meter} meters and \SI{1.14}{\meter} in the mid-range setting. Similarly, 'Camera-With-GNSS' and 'Baseline-With-GNSS' follow the same trend, with sub-meter accuracy and very low standard deviations, indicating highly stable performance.

In summary, while all methods benefit from the increased sensing range of long-range LiDAR, the degree of improvement varies. SLAM-based methods like KISS-SLAM see the largest relative gains, though they remain less reliable than simpler or GNSS-supported approaches.

The resulting trajectories are illustrated in \cref{fig:long_range_trajectories} and \cref{fig:mid_range_trajectories}. Across all tested methods, challenges in mapping accuracy become evident, particularly at the outer regions of the environment where loop closures are infrequent. This limitation is most apparent along the road segment in the middle of the left side of the map. This section consists of a straight road with no intersecting paths, limiting the SLAM system’s ability to perform effective loop closures or re-localization. A similar pattern of increased error is observed in the upper central region of the trajectory, where the environment is dominated by open roads lined with sparse vegetation such as trees.

These areas pose significant challenges for SLAM algorithms due to the lack of distinctive geometric structures. In scenes dominated by repetitive or ambiguous features, such as tree-lined roads without architectural landmarks, the system struggles to extract stable keypoints and landmarks necessary for robust ego-motion estimation and accurate map construction. In contrast, the central area of the environment, which contains a higher density of buildings and other well-defined geometric features, facilitates more reliable feature extraction and data association. Consequently, this region exhibits the lowest localization and mapping errors, highlighting the importance of structured environments for SLAM accuracy.

\begin{table}[th]
\begin{tabularx}{\columnwidth}{r|ccccc}
\hline
\textbf{Mid-Range-Lidar} & \textbf{max} & \textbf{mean} & \textbf{min} & \textbf{rmse} & \textbf{std} \\ \hline
\textbf{KISS-SLAM}      & 15.65 & 5.05  & 0.02 & 5.95 & 3.15 \\
\textbf{Baseline}       &  8.35 & 1.80  & 0.00 & 2.12 & 1.11 \\
\textbf{Camera-No-Ground}     & 9.06  & 1.92  & 0.02 & 2.26 & 1.19 \\
\textbf{Camera-With-Ground}     & 11.40 & 1.95  & 0.02 & 2.35 & 1.32 \\
\textbf{LiDAR-Non-Ground}     & 11.51     & 1.96     & 0.01    & 2.38    & 1.35    \\
\textbf{LiDAR-With-Ground}     & 10.38     & 1.91     & 0.00    & 2.31    & 1.30    \\
\textbf{Baseline-With-GPS}   & 3.70  & 1.00  & 0.00 & 1.16 & 0.58 \\
\textbf{Camera-With-GPS}   & 3.65  & 1.00  & 0.00 & 1.16 & 0.58 \\
\textbf{LiDAR-With-GPS}   & 3.64     & 0.99     & 0.00    & 1.14    & 0.56    \\
\hline
\end{tabularx}
\caption{Quantitative comparison of different localization methods using the mid-range LiDAR. Metrics include maximum, mean, minimum, RMSE, and standard deviation of localization error (in meters). Lower values indicate better performance.}
\label{tab:mid-range}
\end{table}

\section{CONCLUSION}
\label{sec:conclusion}

This study highlights how localization performance improves with long-range LiDAR and GNSS integration, but also emphasizes the crucial role of semantic segmentation from both camera and LiDAR data, especially in GNSS-denied settings.

Semantic segmentation enables systems to distinguish between meaningful and less informative features (e.g., filtering out vegetation and ground, while emphasizing buildings and infrastructure). This contextual awareness significantly enhances localization accuracy in areas with poor geometric structure, such as tree-lined roads or open spaces, where traditional SLAM systems often fail.

Segmentation-based methods achieve the lowest errors in structured environments with rich semantic features, particularly when combined with long-range sensing. Integrating semantic information with geometric data leads to more robust, reliable localization, making it a key component for autonomous navigation in complex real-world environments.

\section*{ACKNOWLEDGMENT}

This paper was created in the “Country 2 City - Bridge” project of the German Center for Future Mobility, which is funded by the German Federal Ministry of Transport.

\printbibliography

\end{document}